\documentclass[journal]{IEEEtran}
\usepackage{comment}
\usepackage{multirow}
\usepackage{graphicx}
\usepackage{float}
\usepackage{subcaption}
\usepackage{epstopdf}
\usepackage{amsmath}
\usepackage{amsfonts}
\usepackage{algorithm}
\usepackage{algorithmic}
\usepackage{makecell}
\usepackage{booktabs}
\setcellgapes{3pt}

\setcellgapes{3pt}

\usepackage[justification=centering]{caption}
\usepackage{color}
\usepackage{multirow}
\usepackage[table,xcdraw]{xcolor}
\usepackage{tabularx}

\usepackage{cite}

\ifCLASSINFOpdf
\else
\fi
\begin{document}
\bstctlcite{IEEEexample:BSTcontrol}
%
\title{Large Language Model Enhanced Multi-Agent Systems for 6G Communications}

\author{Feibo Jiang, \textit{ Member, IEEE}, Li Dong, Yubo Peng,  Kezhi Wang, \textit{Senior Member, IEEE}, Kun Yang, \textit{Fellow, IEEE}, Cunhua Pan, \textit{Senior Member, IEEE}, Dusit Niyato, \textit{Fellow, IEEE}, Octavia A. Dobre, \textit{Fellow, IEEE}
}
\markboth{Submitted for Review}%
{Shell \MakeLowercase{\textit{et al.}}: Bare Demo of IEEEtran.cls for IEEE Journals}
%



\maketitle


\begin{abstract}
The rapid development of the Large Language Model (LLM) presents huge opportunities for 6G communications, e.g., network optimization and management by allowing users to input task requirements to LLMs by nature language. However, directly applying native LLMs in 6G encounters various challenges, such as a lack of private communication data and knowledge, limited logical reasoning, evaluation, and refinement abilities.
Integrating LLMs with the capabilities of retrieval, planning, memory, evaluation and reflection in agents can greatly enhance the potential of LLMs for 6G communications. To this end, we propose a multi-agent system with customized communication knowledge and tools for solving communication related tasks using natural language, comprising three components:
(1) Multi-agent Data Retrieval (MDR), which employs the condensate and inference agents to refine and summarize communication knowledge from the knowledge base, expanding the knowledge boundaries of LLMs in 6G communications;
(2) Multi-agent Collaborative Planning (MCP), which utilizes multiple planning agents to generate feasible solutions for the communication related task from different perspectives based on the retrieved knowledge;
(3) Multi-agent Evaluation and Reflecxion (MER), which utilizes the evaluation agent to assess the solutions, and applies the reflexion agent and refinement agent to provide improvement suggestions for current solutions. 
Finally, we validate the effectiveness of the proposed multi-agent system by designing a semantic communication system, as a case study of 6G communications.

\end{abstract}


\IEEEpeerreviewmaketitle

\section{Introduction}

The future generation of wireless communication, e.g., 6G, is anticipated to provide exceptional data rates, ultra-low latency, and significantly enhanced capacity to accommodate a massive number of user devices. To support the above vision, several innovative techniques such as edge intelligence and Semantic Communication (SC) have been proposed, where Artificial Intelligence (AI)/Machine Learning (ML) has been applied as a key enabling technology. However, the current intelligent communication system design is mainly based on the traditional AI/ML that can be seen as a discriminative AI technique that faces the following challenges, when applied in 6G communications.




\subsection{Challenges of discriminative AI for 6G}

\subsubsection{For dynamic environments}

Future communication systems are expected to operate in rapidly changing environments, due to a variety of factors, such as the movement of devices and network traffic fluctuations. However, the traditional discriminative AI/ML mainly relies on learning local features, leading to trapping local extreme values or having difficulties in learning the long-term dependency of the dynamic network as well as achieving stable operation in a scalable way. Large AI Models (LAMs), as state-of-the-art pretrained foundation models, utilizing multi-head attention mechanisms with even trillions of parameters, enable capturing of a large number of features from a global perspective, which allows the system to achieve a global optimal solution effectively, regardless of how the system changes.

 \subsubsection{For heterogeneous devices} 
 Future communication systems will support a variety of devices, e.g., Internet of Things (IoT) or Unnamed Aerial Vehicles (UAVs), as well as provide various management strategies, like beamforming design, user association, and edge resource allocation. However, the traditional discriminative AI/ML is mainly based on learning task specific features, e.g., only focusing on one type of task. However, LAMs, on the other hand, trained on various types of data and tasks, can be seen as universal models, allowing the same trained model to address different kinds of tasks, e.g., through prompting or fine-tuning processes.
 
\subsubsection{For different applications} 

Future communication systems  need to provide customized solutions for different application scenarios, such as Virtual Reality (VR) and Augmented Reality (AR). For example, in autonomous driving services, the system requires extremely low latency and high reliability transmission, whereas in IoT applications, it must support a massive number of connections. Traditional discriminative AI/ML consists of small models trained for specific application scenarios, limiting them to those particular contexts. In contrast, LAMs possess astounding understanding and creativity and can comprehend and adapt to various application scenarios, allowing them to provide personalized services for different applications.

\subsection{Opportunities of generative AI for 6G}

LAMs offer an entirely new paradigm for solving the above-mentioned challenges \cite{chen2023big}. LAMs represent a significant advancement in generative AI, leveraging their immense size, extensive computational requirements, and vast amounts of training data to achieve state-of-the-art performance in various tasks. 
Their ability to understand intent and generate solutions opens up new possibilities for improving a wide range of applications for 6G communications. LAMs have the potential to revolutionize how we interact with and utilize AI in networks. The main features of LAMs are as follows:

\subsubsection{Multi-head self-attention} Multi-head self-attention allows LAMs to focus on the global perspective and it can analyze and capture spatio-temporal dependencies in changing environments at different scales. This mechanism enables the LAM to generate stable and timely responses, unlike traditional recurrent neural networks that require retraining to adapt to environmental changes.
For example, the multi-head attention enables comprehensive learning of dynamic factors in the network, such as user mobility and traffic fluctuations. This mechanism avoids the long-term forgetting effect caused by dynamic environments, leading to accurate traffic prediction and optimal resource allocation.

\subsubsection{Universal task model} LAMs typically have an extensive number of parameters, which can range from tens of billions to trillions. The large number of parameters allows LAMs to capture intricate network patterns and nuances between heterogeneous devices and imbalance data during training. 
For example, by learning the Channel State Information (CSI), the constraints for computation, communication, and storage resources of various edge devices and edge servers, it is possible to design a universal offloading model that achieves offloading optimization and resource scheduling for different system models or optimization objectives using prompts, without the need for retraining the model.

\subsubsection{Astounding understanding and creativity} LAMs have demonstrated remarkable abilities that go beyond analyzing and generating human language. These abilities stem from the vast amount of knowledge and patterns they acquire during training. 
Based on its exceptional understanding capabilities, LAM can proactively analyze user demands and preferences in 6G networks, enabling the provision of personalized computing and communication services. Leveraging its astonishing creativity, LAM can dynamically plan, configure, and optimize the future communication network through self-learning and self-adaptation abilities.

\subsection{Contributions}
In the article, 
we describe the possible roles of Large Language Models (LLMs) and how to unleash their potential in future communication networks. To overcome the current challenges of applying LLMs to 6G, we propose an LLM-enhanced multi-agent system with customized communication knowledge and tools, which leverages collaboration and interaction among multiple agents to optimize the task-solving capabilities in 6G networks. Specifically, users express their task requirements by natural language firstly. Then, the Multi-agent Data Retrieval (MDR) is proposed to query and summarize domain-specific knowledge in 6G communications from private data. Next, a novel Multi-agent Collaborative Planning (MCP) decomposes the original task based on retrieved communication knowledge, generates multiple feasible sub-task chains, and solves them. Subsequently, the Multi-agent Evaluation and Reflexion (MER) is proposed to evaluate, reflect, and improve the current feasible solutions. As a whole, these form a self-learning and adaptive multi-agent system for solving communication-related problems by natural language. Finally, we validate the effectiveness of the multi-agent system through a case study.


\section{How LLMs Support 6G Communications}
LLMs constitute the most significant category of LAMs. On one hand, LLMs exhibit stronger comprehension, decision-making, and robustness compared to traditional ML models widely used in wireless networks. This higher level of intelligence brings new opportunities for sensor, communication, and computation in 6G wireless networks, enabling efficient and scalable general-purpose wireless intelligence. On the other hand, the massive and diverse wireless data, along with ubiquitous wireless devices in 6G wireless networks, provide powerful support in terms of data and computational resources for LLMs \cite{chen2023big}. Therefore, the application of LLMs to enhance the intrinsic intelligence of 6G networks holds significant importance.
\subsection{The roles of LLMs in 6G communications}
In 6G communications, LLMs can play the following roles and fulfill functions as:
\subsubsection{Data generator}

LLM is a powerful generative AI, which can generate specific types of data based on their domain knowledge. Some novel generative structures (e.g., autoregression decoder and diffusion model) are introduced to LLMs for creating data efficiently. For instance, LLMs can generate high-quality synthetic CSI data without identification information for network optimization, encompassing aspects such as positioning, bandwidth allocation, and network architecture design. This data can assist operators in more effectively planning and developing their 6G networks without violating privacy.

\subsubsection{Knowledge organizer}
LLMs can reprocess or mine raw data for knowledge extraction and analysis, which take both user requirements and raw data processed as input and utilize the extensive domain knowledge of LLMs to deduce new information. For example, LLMs can be introduced as the knowledge base to assist the encoder of SC and can reduce ambiguity and enhance semantic understandings \cite{jiang2023large}. 
\subsubsection{Task scheduler}
LLMs can understand instructions and schedule algorithms or protocols to collectively address complicated communication tasks. By using LLMs as a bridge between communication requirement and solutions, LLMs can manage and invoke proper algorithms, and collaboratively fulfill task requirements while producing results. For instance, LLMs can autonomously allocate service areas to different UAVs, plan their trajectories, guide UAVs to avoid obstacles, and provide critical communication links and computational resources in emergency environment \cite{zhong2023safer}.

\subsubsection{System designer}

LLMs possess powerful natural language understanding and deduction capabilities, enabling them to design communication systems based on given system requirements. They exhibit strong multi-task processing and multi-module design capabilities within the boundaries of various training data. For instance, utilizing its intrinsic AI knowledge, LLMs can automatically design a federated learning system based on a given functional description of fedavg algorithm, and continuously optimize the system through prompt engineering \cite{shen2023large}.

\subsection{How to unleash the LLM potential for 6G communications}
We can unleash the potential of LLMs in 6G communications by the prompt engineering which employs the following approaches:

\subsubsection{In-context learning}

In-Context Learning (ICL) is a form of analogical learning that incorporates explicit examples in the prompt to assist LLMs in making decisions \cite{chan2022data}. It enables the LLM to generate accurate expected results for a new task based on one provided example (i.e., one shot learning) or few similar examples (i.e., few shot learning) without the need for fine-tuning model weights. For instance, ICL can be applied to anomaly detection in industrial IoT. By providing few anomaly examples in the prompt, the LLM can learn and adapt to unique patterns and behaviors of edge devices, enabling real-time detection of anomalies and potential faults without relying on model training.


\subsubsection{Chain-of-Thought}
Chain-of-Thought (CoT) is a form of discrete prompt learning that goes beyond providing examples with input-output pairs \cite{wei2022chain}. It also includes the thought process and steps leading to the desired output when presenting the examples. This approach guides the LLMs’ way of thinking in their reasoning process step-by-step, thereby enhancing the logical reasoning capabilities of the LLMs. For instance, in trajectory planning of UAVs for 6G communications, CoT can be used to break down the trajectory planning process into multiple stages. Each stage focuses on a specific aspect, such as path generation, obstacle avoidance, or mission objectives. The outputs from one stage serve as inputs to the next, ensuring a coherent and sequential planning process.

\subsection{Challenges of applying LLM in 6G  communications}
Challenges arise when applying LLMs to 6G, and they can be categorized as follows:

\subsubsection{Untimely and covertly private data}

LLMs are typically trained on static datasets, while data and information in 6G communication may constantly change. Especially, a vast amount of data can be generated daily at the edge, and a significant portion of this data comprises private and confidential information, rendering it inaccessible for public training purposes. As a result, the LLMs may not capture the latest data and concepts, emerging protocols and standards in a timely manner. This can lead to outputs that are less accurate or not fully aligned with the current communication system.

\subsubsection{Lack of domain knowledge}

6G communication has specific technical requirements and constraints, such as extremely low latency and very high data rate. However, LLMs may be trained on general data sets, which lack an in-depth understanding of domain-specific expertise. For example, the latest GPT-3.5 does not have an inherent understanding of what the SC system is and its structure. Instead, it interprets the SC system as a mode of semantic interaction employed by humans.
This can result in degraded performance of the LLM in communications, as it may struggle to accurately predict and optimize specific task parameters and system performance.

\subsubsection{Insufficient logical reasoning ability}

6G communication involves complex signal processing, optimization, and decision-making tasks. Although LLMs excel in language generation and comprehension, they have limitations in logical reasoning and inferential capabilities. For instance, tasks like wireless channel estimation or resource scheduling require intricate reasoning and decision-making abilities, where LLMs may struggle to perform accurately. This may result in logical errors, or lack understanding of causal relationships, which is common in optimization tasks in 6G communications.

\subsubsection{Inadequate evaluation and refinement}

The evaluation of outputs from LLMs by the single-instance result can be challenging due to the complexity and diversity of wireless communication environments. Assessing the LLM's performance and effectiveness in real-world scenarios requires considering multiple factors, such as channel conditions, user requirements, and mobility. Therefore, evaluating LLMs necessitates adopting multiple perspectives, considering various factors, and providing feedback to continuously refine the performance of LLMs to ensure reliability and usability.

Hence, we could improve the performance of LLMs in 6G communications by training on up-to-date datasets, incorporating communication knowledge and reasoning capabilities, and developing more comprehensive evaluation and refinement methods.

\section{LLM Enhanced Multi-Agent Systems}
\subsection{LLM enhanced agent system}

\begin{figure}[htpb]
	\centering
	\includegraphics[width=9cm]{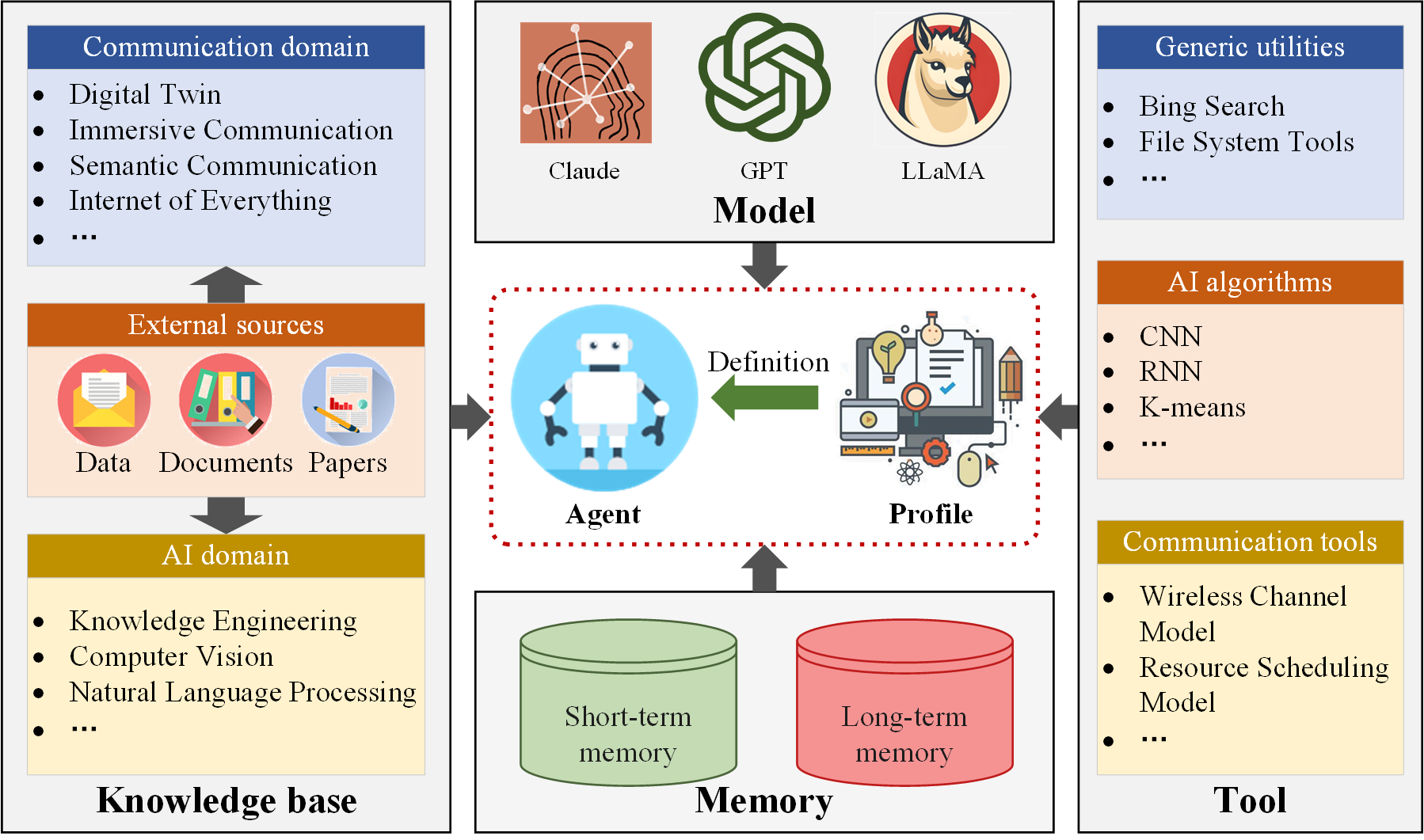}
	\caption{LLM enhanced agent system.}
	\label{fig:agent}
\end{figure}

AI agents can address the above challenges, and they are computational entities designed to simulate or mimic intelligent behavior, possessing traits such as autonomy, reactivity, and communication capacity. Due to their versatile and remarkable capabilities, LLMs are regarded as powerful tools for constructing AI agents. In the paper, we define a typical LLM enhanced agent system in Fig. \ref{fig:agent}, which has the following components:

\begin{itemize}	
	
	\item \textbf{Knowledge base} represents the component that stores the latest private data, consisting of communication standards, documents, and papers that can be connected to the agent, and provides a way to index, query, and update domain-specific knowledge in 6G communications.  
	
	\item \textbf{Tools} are interfaces that an agent can use to process tasks. They can be generic utilities (such as Bing search and file system tools), endogenous intelligence in the LLM (such as AI model), or customized communication models (such as the wireless channel model). Tools can also be defined by the user, or integrated from external sources. 
	
	\item \textbf{Memory} is the component that stores the intermediate and final outputs of agents for evaluation and reflexion, providing a way to manage, retrieve, and modify these historical outputs. 
	
	\item \textbf{Model} is introduced to comprehend and interpret human language inputs, extracting meaning, intent, and context. Agent system supports different LLMs such as OpenAI's GPT, Meta's LLaMA and Anthropic's Claude.

	\item \textbf{Agent} represents the entity that interacts with LLMs, memories, tools and knowledge base, providing a way to plan, control, introspect and communicate with other agents using LLM with a profile. The profile defines and manages the characteristics and behaviors of an agent by specific prompt. It encompasses a set of parameters and rules that describe various attributes of the agent, including its role, goals, capabilities, knowledge, and behavioral patterns.
\end{itemize}

\subsection{Overview of the proposed multi-agent system}

We design an LLM enhanced multi-agent system for 6G communications, which constructs a specialized knowledge base and tools for 6G communications, and possesses planning, memory, tool utilization and introspection capabilities beyond protogenetic LLMs.
Here, an agent serves as the core of this system, which can engage in planning and reflexion based on LLMs, acquire specialized knowledge and tools, and leverage their combination for autonomous learning and adaptive enhancement \cite{wu2023autogen}. 
Moreover, to avoid biases and hallucinations caused by a single agent, we introduce the multi-agent collaboration, which involves designing multiple agents to engage in multi-round cooperation. By combining their individual opinions and knowledge, this system enhances the problem-solving capabilities in complex communication tasks and fully unleashes the cognitive synergy potential of the LLM. 

\begin{figure*}[htpb]
	\centering
	\includegraphics[width=18cm]{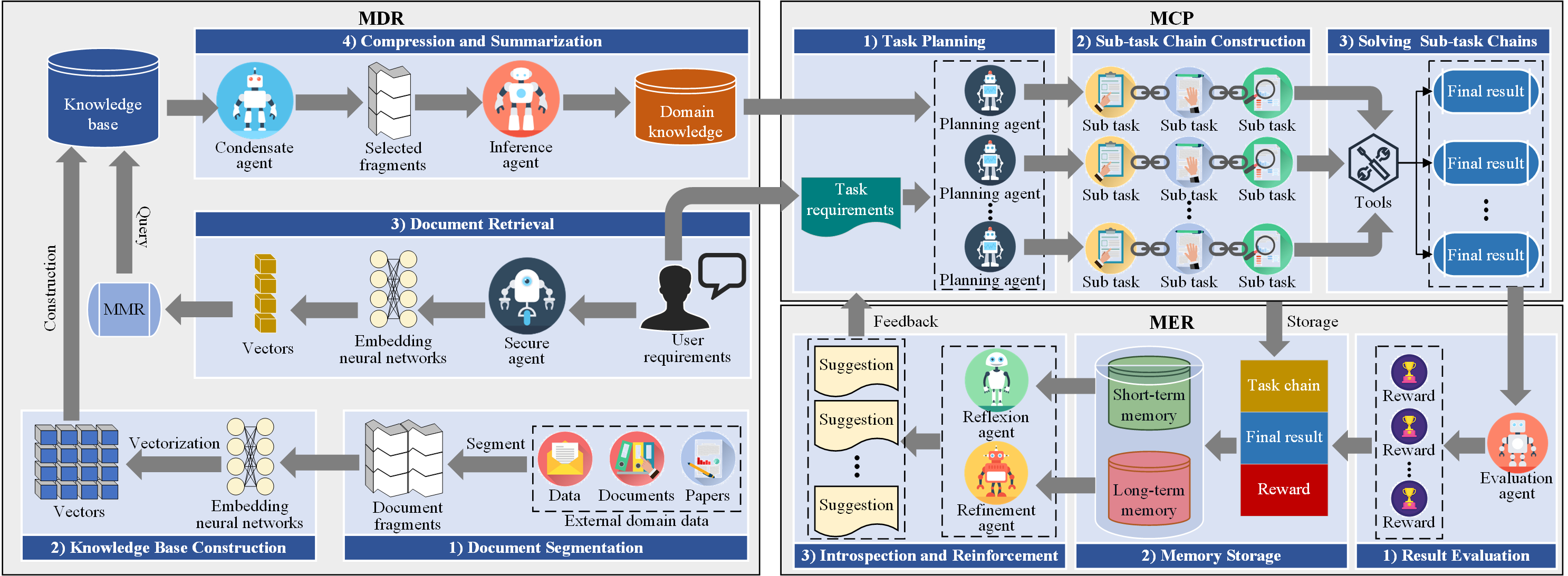}
	\caption{The proposed LLM enhanced multi-agent system.}
	\label{fig:MGPT}
\end{figure*}

As depicted in Fig. \ref{fig:MGPT}, the process begins with the MDR module querying communication knowledge from external data sources based on user requirements by natural language. This step involves several agents, including a \textit{secure agent}, a \textit{condensate agent}, and an \textit{inference agent}.
Once the communication knowledge is obtained along with task requirements, they are fed into the MCP module. Then, multiple \textit{planning agents} and sub-task chains are employed to formulate solutions for the given task and retrieved knowledge. With assistance from the tools, we can derive the final results based on solving the sub-task chain.
Subsequently, the MER module introduces an \textit{evaluation agent} that assesses the final results of each sub-task chain and assigns corresponding rewards. Furthermore, it uses a \textit{reflexion agent} and a \textit{refinement agent} to provide refined suggestions.
Finally, these suggestions are looped back to the MER module, guiding it to re-plan and generate new sub-task chains and results. By iterating through this process, we eventually reach an optimal solution that is then delivered to the user by natural language.

\subsection{Multi-agent data retrieval}
MDR enables LLMs to extract and summarize knowledge from external privacy data sources regarding the 6G communications. 
MDR can also use the LLM as a reasoning engine over new domain knowledge provided in the knowledge base. The steps of MDR are as follows:
\subsubsection{Document segmentation}
External domain data, including latest communication standards, documents, and papers in various formats, can be loaded and then segmented to improve the efficiency of data retrieval. The segmentation process involves dividing the documents into multiple coherent and meaningful fragments while maintaining the semantic coherence and integrity.

\subsubsection{Knowledge base construction}
The segmented document fragments are transformed into numerical vectors using embedding neural networks. Document fragments with similar semantic content will have similar vectors in the numerical space. By comparing these vectors, we can identify similar text fragments. The segmented document fragments and their corresponding embeddings are stored in vector format to construct a knowledge base, facilitating subsequent retrievals.

\subsubsection{Document retrieval}
A \textit{secure agent} is employed to scrutinize user requirements, thereby preventing any unauthorized requests or potential injection attacks \cite{topsakal2023creating}. Once validated, these legal requirements are transformed into vectors through the use of embedding networks. These requirement vectors are then compared with document vectors already stored in the knowledge base, and the most closely matching vectors along with their corresponding document fragments are selected. To increase retrieval diversity and minimize redundancy during this querying process in the knowledge base, we employ the Maximum Marginal Relevance (MMR) \cite{luan2018mptr} for selecting document fragments.

\subsubsection{Compression and summarization}
To reduce irrelevant information in the documents, a \emph{condensate agent} is utilized to compress the documents, resulting in more accurate and focused results \cite{topsakal2023creating}. The selected fragments, along with the query, are then inputted into an \emph{inference agent} to obtain specialized communication knowledge corresponding to the user's requirements by natural language \cite{wei2022chain}.

\subsection{Multi-agent collaborative planning}
MCP generates multiple feasible sub-task chains by constructing multiple \emph{planning agents}. By combining their individual knowledge and planning capabilities from different perspectives, the quality of response in solving complex problems is enhanced. The steps of MCP are as follows:
\subsubsection{Task planning}
 Based on the current task requirements and retrieved communication knowledge, multiple \emph{planning agents} are initialized. Each agent employs either the CoT or Plan-and-Solve approach \cite{wang2023plan} to decompose the original task into a series of sub-tasks.

\subsubsection{Sub-task chain construction}
Considering the order and dependency relationships of all sub-tasks, a sub-task chain is constructed by connecting the individual sub-tasks in a sequential or parallel manner, ensuring the coherence and uniformity of all sub-tasks.

\subsubsection{Solving sub-task chains}
 Each sub-task chain is then solved by invoking either the intrinsic general-purpose tools or external custom tools to address each sub-task separately, until the final result of the sub-task chain is obtained.
 
 \subsection{Multi-agent evaluation and reflecxion}
 MER is used to evaluate the quality of the results generated by MCP and then to reflect on the planning results by memory, thereby facilitating automatic learning, continuous improvement, and self-refinement. The steps of MER are as follows:
 
 \subsubsection{Result Evaluation}
All sub-task chains and their results from MCP are collected and the rewards of all planning results are calculated using an \textit{evaluation agent}.
 
 \subsubsection{Memory Storage} 
 A comparison is made between the current sub-task chain and historical sub-task chains. Task chains with significant differences in semantic space are stored in the long-term memory, while task chains with similar semantics are stored in the short-term memory, along with their corresponding results and rewards.

 \subsubsection{Introspection} 
 
A \emph{reflexion agent} is employed to extract the fine-grained information from the short-term memory, similar to how humans can recall recent details \cite{shinn2023reflexion}. It allows for contemplating the performance of the current sub-task chains across historical schemes, providing valuable small-scale feedback for refinement.
 
\subsubsection{Refinement}  
A \emph{refinement agent} is utilized to reference coarse-grained information from the long-term memory, similar to how humans extract important experiences from long-term decisions \cite{madaan2023self}. It involves contemplating the performance of the current sub-task chains from a global perspective and providing large-scale feedback for improving the sub-task chains. 
  
The introspection and refinement enable us to examine the contents of sub-task chains from different scales, gaining an in-depth understanding of the effectiveness and applicability of each sub-task.
Based on the feedback of the suggestions proposed by introspection and refinement, the MCP is re-driven to generate new sub-task chains, which are then evaluated through MER. This iterative process continues until the optimal scheme is obtained. All agents used in the system are summarized in Table \ref{tab:agents}.

\begin{table}[htpb]
	\centering\makegapedcells
	\caption{Summary of agents.}
	\label{tab:agents}
	\begin{tabular}{|p{50pt}<{\centering}|p{140pt}<{\centering}|p{15pt}<{\centering}|}
		\hline
		Agent               & Function & Ref.\\ \hline
		Secure agent        &    Prevents any unauthorized requests or potential injection attacks       & \cite{topsakal2023creating}           \\ \hline
		Condensate agent    &   Compresses retrieved documents for more accurate and focused results       &      \cite{topsakal2023creating}      \\ \hline
		Inference agent     &    Concludes specialized domain knowledge corresponding to user's requirements      &        \cite{wei2022chain}       \\ \hline
		Planning agent      &      Decomposes the original task into a series of sub-tasks    &\cite{wang2023plan}          \\ \hline
		Evaluation agent    &   
		Evaluates results of sub-task chains and calculate the rewards of planning agents     &      -     \\ \hline
		Reflexion agent     &    Extracts the fine-grained information from the short-term memory      &   \cite{shinn2023reflexion}         \\ \hline
		Refinement agent &   References coarse-grained information from the long-term memory       &     \cite{madaan2023self}       \\ \hline
	\end{tabular}
\end{table}

\section{Case Study}
\begin{figure*}[htpb]
	\centering
	\includegraphics[width=18cm]{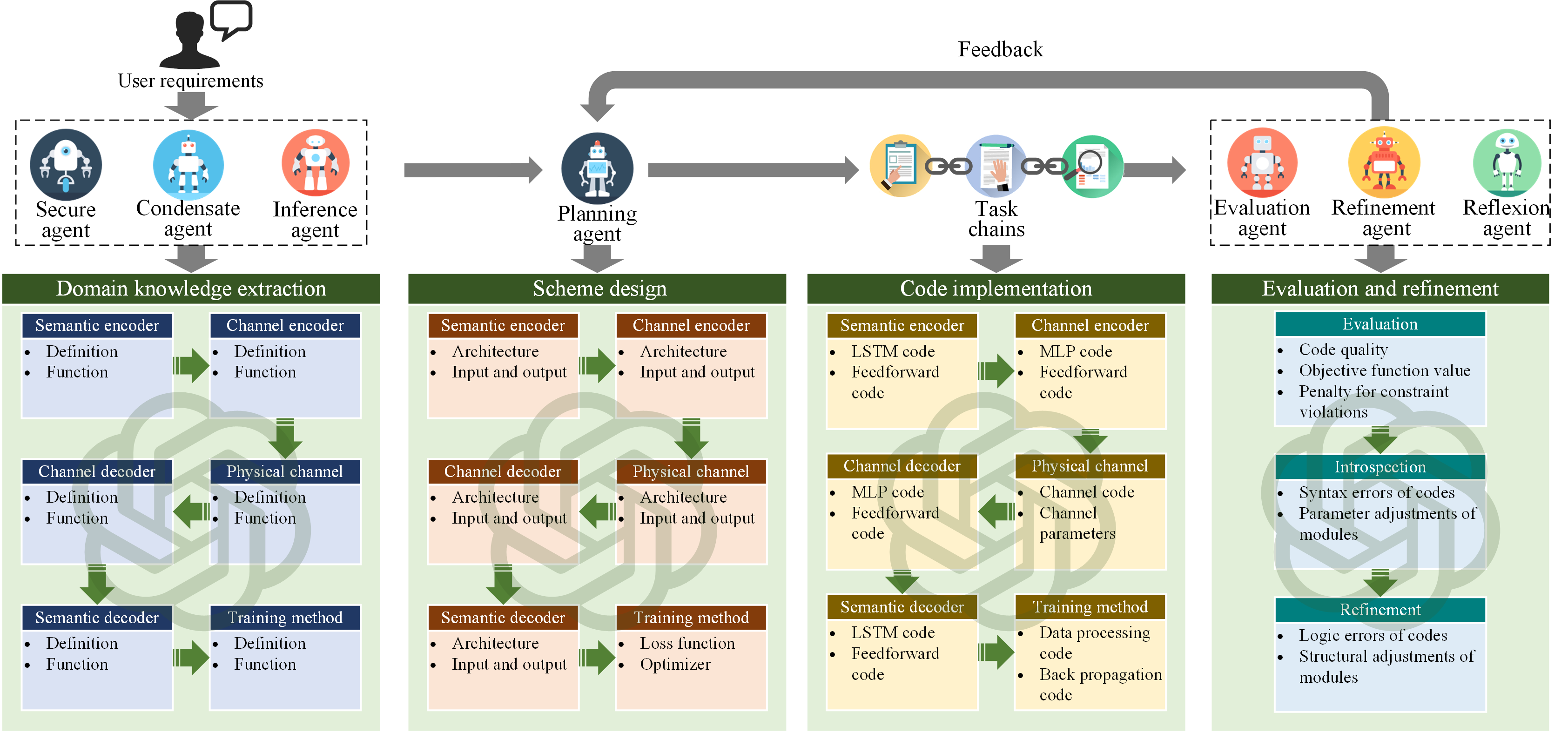}
	\caption{The illustration of implementing the SC model based on the LLM-enhanced multi-agent system.}
	\label{fig:SC}
\end{figure*}

In this section, we introduce a case study in 6G communications, i.e., LLM for SC, which is expected to become the key applications
of next generation intelligent communication paradigms. Specifically, we utilize the proposed multi-agent system to automatically generate an SC model based on user requirements and resource constraints.

\subsection{SC model designed by the multi-agent system}
As illustrated in Fig. \ref{fig:SC}, we apply the multi-agent system to construct the SC model. In this study, we chose GPT-3.5 as our LLM. The input content is expressed in natural language, encompassing user requirements, design objectives and constraints, as well as evaluation metrics as follows:

\textbf{Input}:
\emph{Please produce Python code that implements an SC model for text transmission. 
The Additive Gaussian White Noise (AWGN) can serve as the physical channel. 
The bilingual evaluation understudy (BLEU) score is adopted as the metric that evaluates the SC model.
We expect that the generated SC model achieves no less than a 0.6 BLEU score when the Signal-to-Noise Ratio (SNR) is 10 dB. The total number of model parameters does not exceed 2,000,000 for the resource constraints of devices.}

It is worth noting that the current multi-agent system lacks any knowledge about the SC system beyond the input, as the native GPT-3.5 does not understand the structure and functionality of SC. The multi-agent system needs to autonomously learn from the knowledge base and address these issues. The output is the outline and implementation code for this SC model. In this study, the specific contents of the outline part are listed as follows:

\textbf{Output}: \emph{The Python-based SC model has been successfully implemented, incorporating all necessary modules. The semantic encoder and decoder are realized using Long Short-Term Memory (LSTM) networks. The channel encoder and decoder are constructed based on the Multilayer Perceptron (MLP) architecture.  The final SC model can achieve a 0.68 BLEU score when SNR is 10 dB, which meets expectations. In addition, the total number of model parameters is 1,826,762.}


We have gathered a selection of research papers from arxiv that are relevant to 6G communications to build a basic communication knowledge base. In addition, we have included essential communication models, such as channel models, as specialized communication tools. 
Next, the specific progress of applying the multi-agent system to design the  SC model is as follows:
\begin{enumerate}	
	\item The \emph{secure agent} is utilized to validate the legitimacy of the input, while the relevant arxiv papers (e.g., \cite{yang2022semantic}) in the knowledge base are retrieved. Then, the \emph{condensate agent} is applied to refine the retrieved SC knowledge in papers. Next, the \emph{inference agent} amalgamates user input and SC knowledge to distill relevant SC knowledge for constructing the SC model, which includes the definitions and functions of each module in the SC.

	\item Once the necessary SC knowledge is obtained, each \emph{planning agent} formulates a specific sub-task chain for the SC model. This chain delineates the architecture of each module in the SC model, including their respective inputs and outputs, network structure, as well as training settings (e.g., loss function and optimizer).

	\item Sub-task chains are then handled by the endogenous AI code generation tool of the LLM. Each sub-task chain manages one feasible scheme of the SC model including semantic encoder/decoder, channel encoder/decoder, and other codes (e.g., data processing, feedforward and back propagation). The wireless channel code is generated by the predefined communication tools (e.g., channel models). This results in obtaining precisely Python code.

	\item The \emph{evaluation agent} subsequently assesses the quality of generated codes.
	The evaluative score encompasses three components: the quality of the generated code, the value of the objective function, and the penalty for constraint violations.
The value of the objective function is defined as the BLEU score of the designed SC model.

\item Based on these evaluation results, the \emph{reflexion agent} gives fine-grained introspective comments (e.g., syntax errors of codes and parameter adjustments of modules), and the \emph{refinement agent} gives coarse-gained reinforced comments (e.g., logic errors of codes and structural adjustments of modules) taking into account both strengths and weaknesses of the current SC model.

	\item These proposed improvement suggestions are then fed back into the planning agent for introspection and refinement purposes. This process undergoes iterative optimization until satisfactory results are obtained.
\end{enumerate}


\subsection{Simulation results}
\begin{figure}[htpb]
	\centering
	\includegraphics[width=9cm]{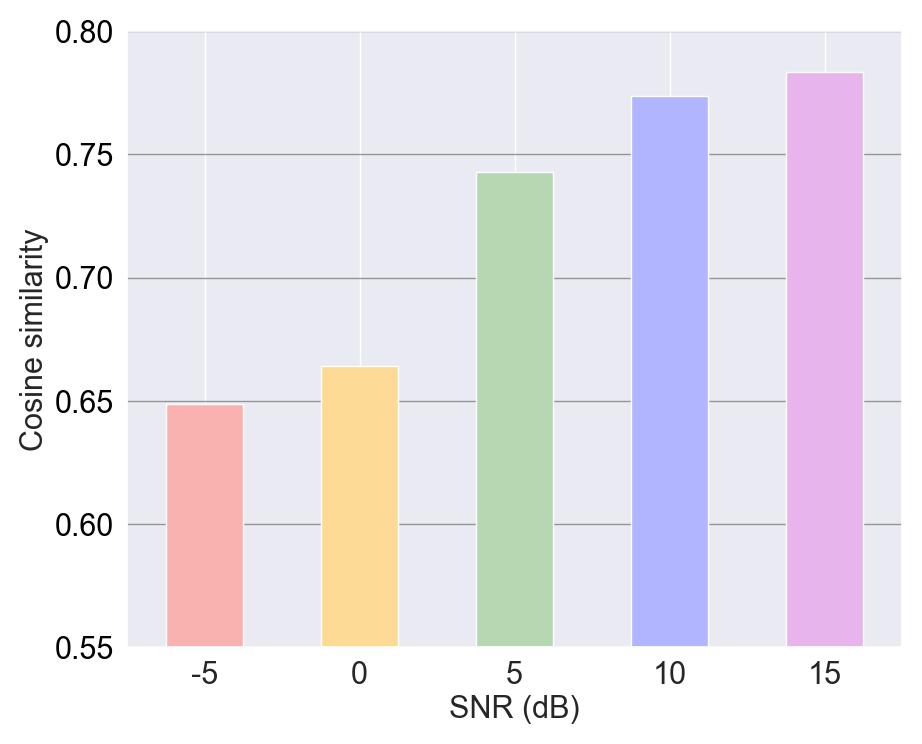}
	\caption{Cosine similarity versus SNR.}
	\label{fig:exp2}
\end{figure}

\renewcommand{\arraystretch}{1.2}
All schemes and experimental codes are generated by GPT-3.5, and all parameter optimizations are performed automatically by the multi-agent system. We only assign two planning agents in the multi-agent system. This implies that two independent SC models, each based on a unique scheme, are concurrently generated. We also set the number of iterations to four.
Fig. \ref{fig:exp2} presents simulation results for the SC model with different structures. It displays evaluative scores for both schemes. Notably, as the iterations advance, there is a distinct enhancement in evaluative scores. Initially, Scheme 2 trailed behind Scheme 1 in terms of evaluation scores. However, in the second iteration, the semantic encoder-decoder of Scheme 2 underwent coarse-gained refinement and transformed to the LSTM structure. As a result, Scheme 2 eventually caught up and surpassed Scheme 1.
The adopted network architectures in the final SC models of the two schemes are described as follows:
\begin{itemize}
\item \emph{Scheme 1}:
An MLP with 3 layers is employed as the semantic encoder and decoder. The MLP with 2 layers is integrated into the channel encoder and decoder. The SC model is trained using the SGD optimizer.

\item \emph{Scheme 2}:
An LSTM with 4 layers is utilized for the semantic encoder and decoder. Similarly, the  MLP with 2 layers is used for both channel encoding and decoding. Training of this SC model is based on the Adam optimizer.
\end{itemize}

In summary, this simulation demonstrates the proposed multi-agent system's ability to autonomously generate an SC model while iteratively refining it through self-introspection and refinement.

\begin{figure}[htpb]
	\centering
	\includegraphics[width=9cm]{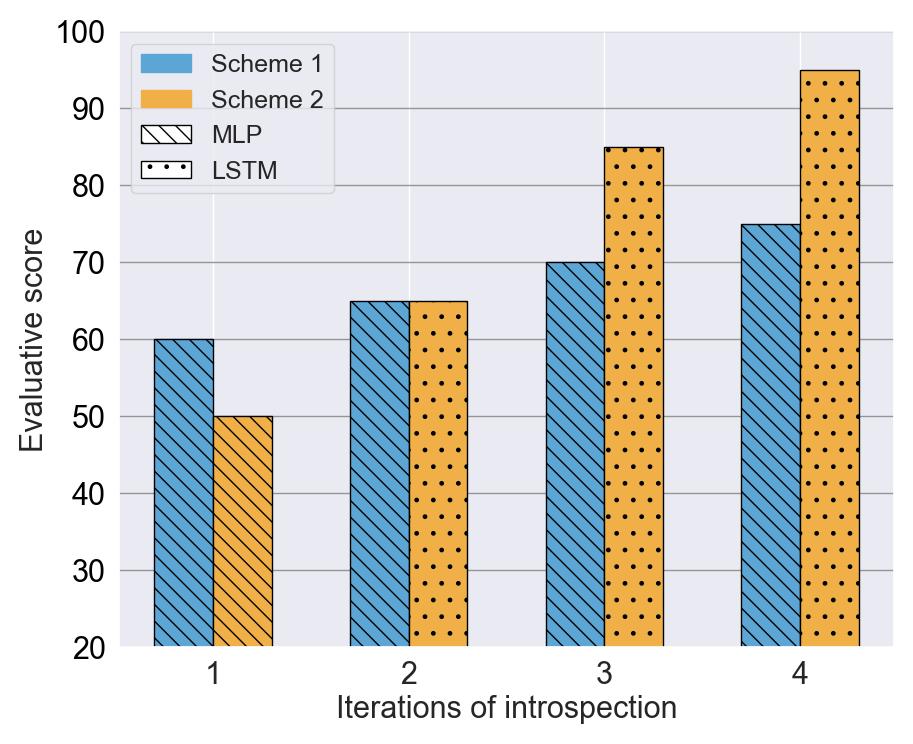}
	\caption{Evaluative score versus iteration number.}
	\label{fig:exp1}
\end{figure}

To assess the effectiveness of the generated SC model by Scheme 2, we utilize the Cornell Movie-Dialogs Corpus dataset \cite{danescu2012you}, a collection of dialogues from 617 film scripts, to train the SC model. Specifically, we use 8,000 dialogues for training and reserve 2,000 dialogues for testing. The training epoch is set at 50.
As for the evaluation metric, we employ a BERT-based semantic evaluation method complemented by cosine similarity \cite{reimers2019sentence}. 
Subsequently, we compute the cosine similarity between these text encodings from raw and recovered text data.
Fig. \ref{fig:exp2} depicts the semantic similarity results of our SC model on the test set while varying the SNR. The figure clearly illustrates that the performance of our SC model improves as the SNR increases. These findings not only showcase the functionalities, but also underscore the effectiveness of the SC model generated through the proposed multi-agent system.

\section{Open Issues}

\subsubsection{Limited Resources}
The multi-agent system relies heavily on the availability of LLMs and the private communication data at the edge. 
However, edge devices often have limited computing, storage and energy resources compared to powerful cloud servers. LLMs are resource-intensive models that may exceed the processing capabilities of edge devices, making it challenging to deploy multi-agent systems on the edge.


\subsubsection{Cooperation and competition} 
The proposed multi-agent system adopts a cooperative approach for all agents to accomplish the design of the SC system. The interaction mode among agents is important for the multi-agent system and LLMs facilitate diverse modes of interaction among agents. 
Exploring alternative modes of interaction, such as competition, and their application in 6G communications would be an intriguing research direction. 
 
  

\subsubsection{Real-time interaction} LLMs often suffer from slow response times, making them impractical for real-time, interactive 6G applications. Developing efficient and faster inference methods to enable real-time interaction with LLM-based agents is an ongoing challenge.

\section{Conclusion}
In this paper, we proposed a multi-agent system that utilized natural language to design solutions for 6G communications and provided the case study in SC tasks. The system leveraged multiple LLM-enhanced agents to collaborate, self-learn, self-improve, and effectively solve defined problems in 6G communications. Specifically, we first employed the MDR to query private data in the system and extracted communication knowledge relevant to the task requirements. Next, we utilized the MCP to generate feasible solutions from different perspectives. Subsequently, we employed the MER to evaluate and reflect on the current solutions, provide improvement suggestions, and guide MCP in enhancing the solutions. Through iterations, an optimal solution was obtained. Finally, we demonstrated and validated the effectiveness of the proposed multi-agent system through the SC case study.


%

%




\bibliographystyle{ieeetran}
\bibliography{bare_jrnl_bobo}
\section*{Biographies}
\textbf{Feibo Jiang} (jiangfb@hunnu.edu.cn) received Ph.D. degree from the Central South University, China. He is currently an Associate Professor at Hunan Normal University, China.

\textbf{Li Dong} (Dlj2017@hunnu.edu.cn) received Ph.D. degree from the Central South University, China. She is currently an Associate Professor at Hunan University of Technology and Business, China.

\textbf{Yubo Peng} (pengyubo@hunnu.edu.cn) is currently pursuing the master’s degree with Hunan Normal University, China. 

\textbf{Kezhi Wang} (Kezhi.Wang@brunel.ac.uk) received Ph.D. degree from University of Warwick, U.K. in 2015. Currently he is a Senior Lecturer with the Department of Computer Science, Brunel University London, U.K.

\textbf{Kun Yang} (kunyang@essex.ac.uk) received his PhD from the Department of Electronic \& Electrical Engineering of University College London (UCL), U.K. He is currently a Chair Professor in the School of Computer Science \& Electronic Engineering, University of Essex, U.K.

\textbf{Cunhua Pan} (cpan@seu.edu.cn) received Ph.D. degrees from Southeast University, China, in 2015. 
He is a full professor in Southeast University, China. 

\textbf{Dusit Niyato} (dniyato@ntu.edu.sg) received the Ph.D. degree in electrical and computer engineering from the University of Manitoba in Canada in 2008. He is a professor in the School of Computer Science and Engineering, Nanyang Technological University, 639798 Singapore.

\textbf{Octavia A. Dobre} (odobre@mun.ca) is a professor and Canada Research Chair Tier 1 at Memorial University, Canada. She is a Fellow of the Canadian Academy of Engineering, Fellow of the Engineering Institute of Canada, and elected member of the European Academy of Sciences and Arts. She is the Director of Journals of the IEEE Communications Society.

\end{document}